\documentclass[lettersize,journal]{IEEEtran}
\usepackage{amsmath,amsfonts}
\usepackage{algorithmic}
\usepackage{algorithm}
\usepackage{array}
\usepackage[caption=false,font=normalsize,labelfont=sf,textfont=sf]{subfig}
\usepackage{textcomp}
\usepackage{stfloats}
\usepackage{url}
\usepackage{verbatim}
\usepackage{graphicx}
\usepackage{cite}
\usepackage{epsfig}
\usepackage{graphicx}
\usepackage{amsmath}
\usepackage{amssymb}
\usepackage{multirow}
\usepackage{bbding}
\usepackage{adjustbox}
\usepackage[table,xcdraw]{xcolor}
\usepackage[normalem]{ulem}
\usepackage{cite}
\usepackage{mathrsfs}
\usepackage{color}
\usepackage{diagbox}
\usepackage{amsthm}
\usepackage{booktabs}
\usepackage{algorithm}
\usepackage{algorithmic}
\begin{document}

\title{EWT: Efficient Wavelet-Transformer for \\ Single Image Denoising}
\author{Juncheng Li, Bodong Cheng, Ying Chen, Guangwei Gao, and Tieyong Zeng

\thanks{This work was supported in part by the National Natural Science Foundation
of China under Grants 61972212 and 61833011, the Shanghai Sailing Program under Grant 23YF1412800, and the Natural Science Foundation of Shanghai under Grant 23ZR1422200. (Juncheng Li and Bodong Cheng
contributed equally to this work.)}

\thanks{Juncheng Li is with the School of Communication \& Information Engineering, Shanghai University, Shanghai, China, also with Jiangsu Key Laboratory of Image and Video Understanding for Social Safety, Nanjing University of Science and Technology, Nanjing, China (E-mail: cvjunchengli@gmail.com).}
\thanks{Bodong Cheng is with the School of Computer Science and Technology, Xidian University, Xian, China. (E-mail: bdcheng@stu.xidian.edu.cn)}
\thanks{Y. Chen is with the Department of Cyberspace Security, Beijing Electronic Science \& Technology Institute, Beijing, China. (E-mail:ychen@besti.edu.cn)}
\thanks{Guangwei Gao is with the Institute of Advanced Technology, Nanjing University of Posts and Telecommunications, Nanjing, China, and also with the Provincial Key Laboratory for Computer Information
Processing Technology, Soochow University, Suzhou, China (E-mail: csggao@gmail.com.}
\thanks{Tieyong Zeng is with the Department of Mathematics, The Chinese University of Hong Kong, New Territories, Hong Kong. (E-mail: zeng@math.cuhk.edu.hk)}
}

\markboth{}%
{Shell \MakeLowercase{\textit{et al.}}: A Sample Article Using IEEEtran.cls for IEEE Journals}


\maketitle

\begin{abstract}
Transformer-based image denoising methods have achieved encouraging results in the past year. However, it must uses linear operations to model long-range dependencies, which greatly increases model inference time and consumes GPU storage space. Compared with convolutional neural network-based methods, current Transformer-based image denoising methods cannot achieve a balance between performance improvement and resource consumption. In this paper, we propose an Efficient Wavelet Transformer (EWT) for image denoising. Specifically, we use Discrete Wavelet Transform (DWT) and Inverse Wavelet Transform (IWT) for downsampling and upsampling, respectively. This method can fully preserve the image features while reducing the image resolution, thereby greatly reducing the device resource consumption of the Transformer model. Furthermore, we propose a novel Dual-stream Feature Extraction Block (DFEB) to extract image features at different levels, which can further reduce model inference time and GPU memory usage. Experiments show that our method speeds up the original Transformer by more than \textbf{80\%}, reduces GPU memory usage by more than \textbf{60\%}, and achieves excellent denoising results. All code will be public.
\end{abstract}


\section{Introduction}
\IEEEPARstart{I}{MAGE} denoising is a popular topic in image restoration (IR), which aims to reconstruct a clean image from the noisy one. As the key step in many practical applications, the quality of denoised images will significantly affect the performance of downstream tasks, such as image classification~\cite{liu2019task,luo2017convolutional}, image segmentation~\cite{liu2018collaborative,ding2022srrnet}, target detection~\cite{zhang2021dardet,wu2023mtu}. However, due to the complex noise environment, image denoising is still a challenging inverse problem.

In the past few decades, researchers have made many explorations and attempts on single image denoising (SID). The method of SID can be divided into traditional denoising methods~\cite{jorgensen2011model,im2012tangent,sun2019block,ma2022retinex,chen2011adaptive} and learning-based methods. Among them, traditional methods are usually implemented in an iterative manner, which is inefficient. In addition, manual design is required and the generalization performance is poor. For learning-based methods, the purpose is to learn the mapping between noisy and clean images, thus making the model has denoising ability. Recently, with the wide application of deep learning in various fields and the excellent performance of convolutional neural networks (CNN) in computer vision, many CNN-based methods~\cite{r13,yang2017bm3d,zhang2018ffdnet,zhang2020residual,park2019densely,sheng2022frequency} have been proposed for SID. Most of them use the powerful feature extraction abilities of CNN to extract image features and use various strategies for modeling, which have achieved gratifying results. Recently, with the proposal and wide application of the visual Transformer model, a new research domain has been provided for SID. Facts have proved that the ability of Transformer to extract long-range dependencies of images makes it have better denoising performance than CNN model. Therefore, some representative image restoration Transformer models~\cite{r18,r19,wang2022uformer,zamir2022restormer} have been proposed.

\begin{figure}
    \centering
    \includegraphics[width=9cm]{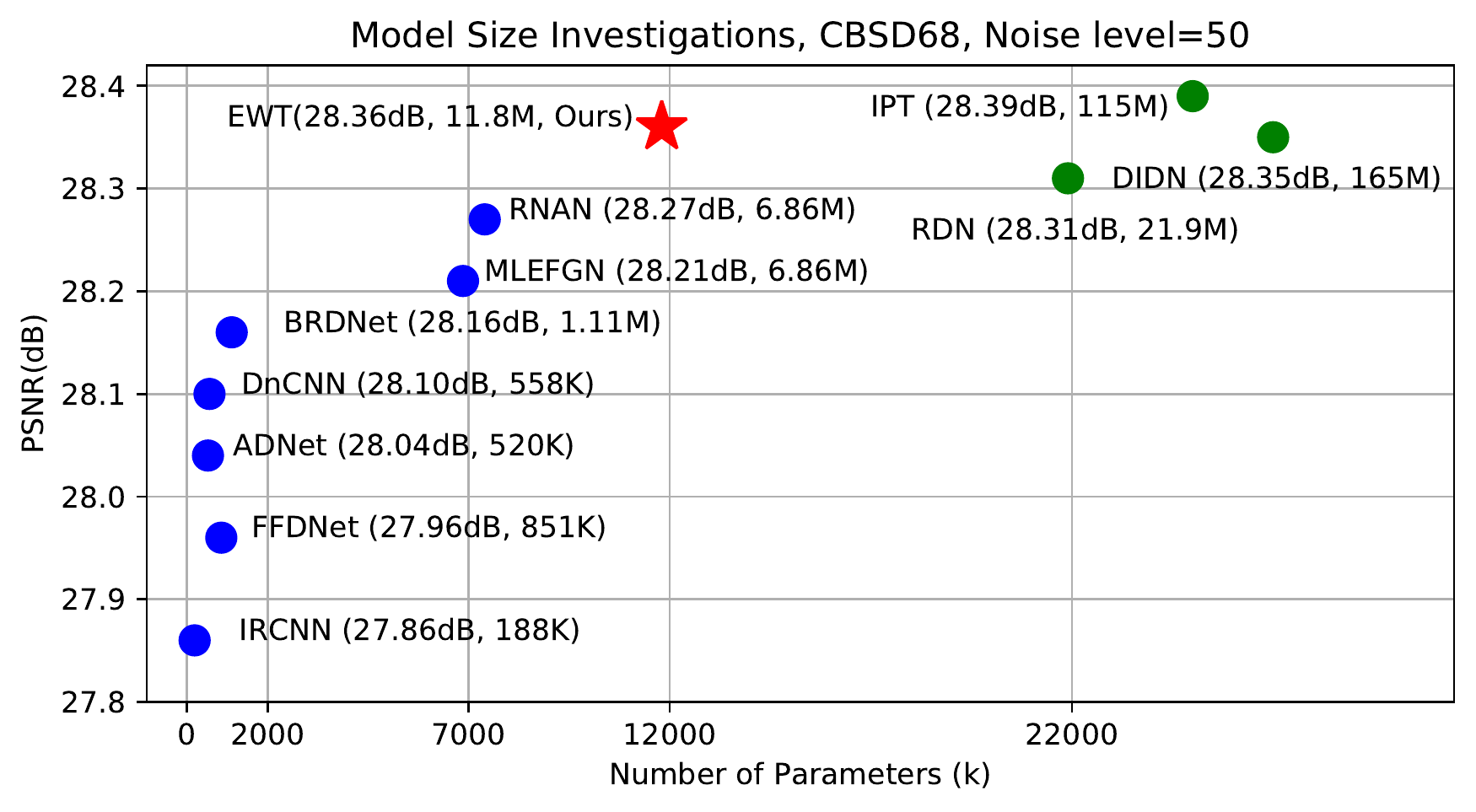}
    \caption{Model performance and size comparison with classic single image denoising methods on CBSD68 ($\sigma$ = 50).}
    \label{Size}
\end{figure}


However, since the mechanism of Transformer is to use matrix operations to operate on the features of each pixel in the image, which will cause excessive consumption of time and space. Although the current Transformer-based image restoration method uses the patch processing method, dividing the image into multiple patches for operation, it still occupy a large amount of GPU memory space, resulting in longer inference time. Therefore, it is difficult to balance the performance and resource consumption of the model. The above problems make
it difficult for the Transformer model to run on server devices with low GPU performance, which greatly limits the research of Transformer on SID tasks.

To overcome the Transformer’s bottleneck in image denoising, we propose a novel Efficient Wavelet Transformer (EWT). Although both DWT and Transformer are common
technologies, as far as we know, this is the first model that introduced the wavelet transform into Transformer and applies it to image restoration task. It is worth mentioning that we do not forcefully combine them but elegantly integrate them according to their own advantages and disadvantages. EWT
uses the reversible nature of wavelets as the sampling unit for model input and output, which can effectively improve the inference speed of the Transformer model and reduce a large amount of GPU memory usage. In the network backbone, we refer to the shift-windows self-attention mechanism in Swin Transformer, and combine the local feature extraction and aggregation capabilities of CNN to construct a dual-stream feature extraction block (DFEB) that combines the respective advantages of Transformer and CNN. In summary, the main contributions of this work are as follows:
\begin{itemize}
  \item We consider the limitations of Transformer in image restoration tasks and propose a novel Efficient Wavelet-Transformer (EWT) for SID. This is the first attempt of Transformer in wavelet domain, which increases the speed of the original Transformer by more than \textbf{80\%} and reduces GPU memory consumption by more than \textbf{60\%}. 
  
  \item We propose an efficient Multi-level Feature Aggregation Module (MFAM). MFAM is a lightweight feature aggregation module that can make full use of hierarchical features by using local and global residual learning. We also propose an elegant Dual-stream Feature Extraction Block (DFEB), which combines the advantages of CNN and Transformer that can take into account the information of different levels to better extract image features.

  \item We fully demonstrate the effectiveness of wavelets in Transformer models. Solve the drawbacks of the slow inference speed and high GPU memory usage of Transformer in image restoration tasks. In other words, EWT is a new attempt to balance model performance and resource consumption, which is helpful for more work in the future.
\end{itemize}

The rest of this paper is organized as follows. Related works are reviewed in Section~\ref{RW}. A detailed explanation of the proposed EWT is given in Section~\ref{Method}. The experimental results, ablation analysis, and discussion are presented in Section~\ref{EXP},~\ref{AS}, and~\ref{DS} respectively. Finally, we draw a conclusion in Section~\ref{CL}.

\section{Related Works}\label{RW}
Recently, several Transformer methods for image denoising have been proposed to demonstrate the effectiveness of the Transformer architecture in this task. Although these methods have achieved good performance, they will occupy a large amount of GPU memory space and prolong the inference time of the network, which is extremely unfavorable for the promotion and application of Transformer in image restoration. In this paper, we aim to explore an efficient Transformer model for image denoising that considers both model performance and resource consumption.

\subsection{CNN-based SID Methods}
With the development of deep learning, CNN-based image restoration methods have achieved advanced results and greatly promoted the development of SID. The success of these methods is attributed to its powerful feature extraction ability and well-designed network structure, which can extract coarse and fine-grained features through different receptive fields. For example, Zhang et al.~\cite{r13} proposed a DnCNN for the Gaussian noise removal, which achieved competitive results by took advantage of batch normalization and residual learning. Yang et al.~\cite{yang2017bm3d} proposed a BM3D-Net, which is a nonlocal-based network that introduced BM3D into CNN by using wavelet shrinkage. Zhang et al.~\cite{zhang2018ffdnet} proposed a flexible FFDNet, which took the noise level map and the noisy image as the inputs for image denoising. Fang et al.~\cite{fang2020multilevel} proposed a multi-level edge features guided MLEFGN, which can make full use of edge features to reconstruct noise-free images. Zhang et al.~\cite{zhang2020residual} proposed an efficient Residual Dense Network (RDN) to extract abundant local features via densely connected convolutional layers. Most of the aforementioned methods committed to budding efficient modules to extract local features to reconstruct noise-free images. In addition, in order to restore more detailed features, many methods~\cite{yu2019deep,park2019densely} directly increase the depth of the network, which results in a substantial increase in the parameters of the model. To better encode image global information, the goal of current research is to explore more powerful deep learning models.

\begin{figure*}[t]
    \centering
    \includegraphics[width=18cm]{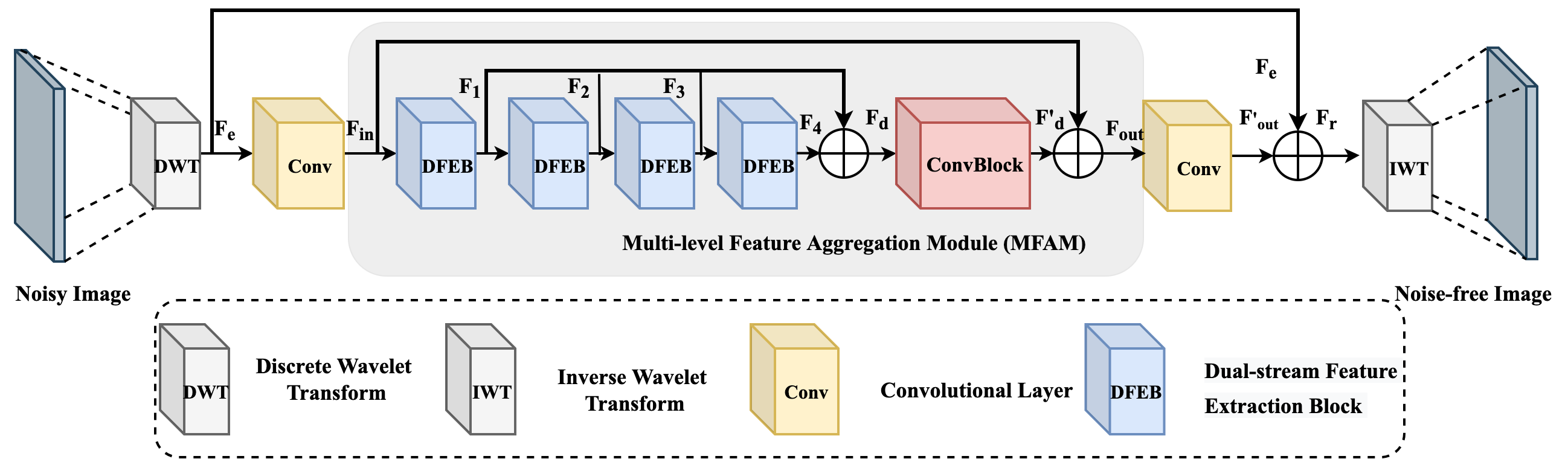}
    \caption{The complete architecture of the proposed Efficient Wavelet-Transformer (EWT), Among them, MFAM is used for feature processing.}
    \label{ewt}
\end{figure*}

\subsection{Transformer-based IR Methods}
In order to model the dependency of pixel-level features, researchers began to pay attention to Transformer in NLP. The self-attention unit in Transformer can well model the long-distance dependencies in the sequence. However, due to the particularity of the image, directly expanding it into a sequence as the input of the Transformer will cause excessive computational overhead. In order to solve this problem, ViT~\cite{r8} uses the idea of dividing an image into multiple sub-images of the same size. Later, in order to better promote the flow of information between sub-images, Swin Transformer~\cite{r11} introduced the idea of window displacement to indirectly model the entire image and demonstrated its excellent performance in high-level vision tasks such as image classification and target detection. Recently, some works also apply Transformer to image restoration tasks, such as IPT~\cite{r18} and SwinIR~\cite{r19}. Among them, IPT draws on the network structure of DERT~\cite{carion2020end}, which use 3$\times$3 convolution with a step size of 3 to reduce the dimensionality of the image. This method can alleviate the dimensionality problem to a certain extent. However, the demanding requirements for GPU memory, training datasets, and reasoning time are unacceptable. SwinIR directly migrated the Swin Transformer to IR task and achieved outstanding results. However, SwinIR stacks a large number of Transformers, the execution time and GPU memory consumption are still very high. Although Transformer can improve the performance of the model, its own mechanism will bring a lot of GPU memory consumption and time overhead. In addition, Transformer cannot encode the two-dimensional position information of the image, and needs to embed relative position or absolute position encoding. In this regard, CNN inherently has the ability to encode the position of the image. Therefore, our goal is to incorporate CNNs and explore a more elegant and efficient Transformer for image restoration.

\subsection{Wavelet-based IR Methods}
Wavelet is widely used in image processing tasks. With the rise of deep learning, some studies combine wavelet with CNN and achieved excellent results. For example, Bae et al.~\cite{r16} found that learning on wavelet sub-bands is more effective, and proposed a Wavelet Residual Network (WavResNet) for image restoration. After that, Bae et al.~\cite{r17} also proposed a deep wavelet super-resolution network to recover the lost details on the wavelet sub-bands. Zhong et al.~\cite{zhong2018joint} jointed the sub-bands learning with CliqueNet~\cite{yang2018convolutional} structures for wavelet domain super-resolution. Liu et al.~\cite{r12} proposed a Multi-level Wavelet-CNN (MWCNN) for image restoration, which use multi-level wavelet to complete related tasks. Inspired by these methods, we intend to explore the performance of Transformer in the wavelet domain and build a more lightweight Transformer model with wavelet.

\section{Efficient Wavelet-Transformer (EWT)}\label{Method}
\subsection{Network Architecture}
As shown in Fig.~\ref{ewt}, EWT mainly consists of three parts: Discrete Wavelet Transform (DWT), feature processing, and Inverse Wavelet Transform (IWT). Specifically, at the top of the model, we first use the DWT to downsample the image, which can effectively extract the high and low-frequency information of the image while reducing the resolution of the image. In the middle part of the model, a Multi-level Feature Aggregation Module (MFAM) is introduced for feature processing. This module can significantly improve the model inference speed while ensuring effective feature extraction. Finally, we use the IWT to restore the image and reconstruct its corresponding noise-free image. Define $I_{noisy}\in H\times W\times C$ as the original input noisy image, the DWT down-sampling layer $f_{DWT}$ will convert $I_{noisy}$ into 4 wavelet sub-images:
\begin{equation}
    I_{LL}, I_{LH}, I_{HL}, I_{HH}= f_{DWT}(I_{noisy}),
\end{equation}
where $I_{LL}, I_{LH}, I_{HL}, I_{HH}\in \frac{H}{2}\times \frac{W}{2}\times C$ are 4 sub-images with different frequencies. We concatenate them as the shallow features $F_{e}\in \frac{H}{2}\times \frac{W}{2}\times 4C$ of EWT, and then use them for feature extraction:
\begin{equation}
    F_{in}= f_{conv}(F_{e}),
\end{equation}
\begin{equation}
    F_{out}=f_{MFAM}(F_{in}),
\end{equation}
where $f_{conv}(\cdot)$ is a 3 $\times$ 3 convolutional layer used to extract the basic information of the image as the initial features. And these features are sent to MFAM to further extract more effective features. After that, a 3 $\times$ 3 convolutional layer also applied on the output $F_{out}$ to obtain the merged features $F^{'}_{out}$: 
\begin{equation}
    F^{'}_{out}= f_{conv}(F_{out}),
\end{equation}
and the global residual learning strategy is used to aggregate $F_{e}$ and $F^{'}_{out}$ as the finally reconstructed feature
\begin{equation}
    F_{r}= F_{e}+ F^{'}_{out}.
\end{equation}

Finally, the IWT operation is used to transform the features to the original resolution and reconstruct the noise-free image
\begin{equation}
    I^{'}_{clean}=f_{IWT}(F_{r}),
\end{equation}
where $f_{IWT}(\cdot)$ denotes inverse wavelet and $I^{'}_{clear}$ is the reconstruct clean image.

During training, EWT is optimized with $L1$ loss function. Given a training dataset $\left \{I_{noisy}^{i}, I_{clean}^{i}  \right \}_{i=1}^{S}$, we solve
\begin{equation}
   \hat{\theta} = \arg\,\min\limits_{\theta}\, \frac{1}{S}\sum_{i=1}^{S}  \left \| F_{\theta}(I_{noisy}^{i}) -  I_{clean}^{i} \right \|_{1},
\end{equation}
where $\theta$ denotes the parameter set of our EWT, $F(I_{noisy})=I^{'}_{clean}$ is the reconstruct noise-free image.

\subsection{Wavelet-based Image Sampling}
Effective sampling of an image is a necessary considered problem in image restoration tasks since the resolution of the input image is usually very large. This means that it will take a lot of calculation costs to deal with them. Although the image size can be reduced by cropping, it will result in the inability to capture the global information of the image. To solve this problem, many methods have been proposed to reduce the image resolution, such as pooling or convolution operations. For image restoration tasks, the final output needs to be restored to the original image size. However, aforementioned operations will cause irreversible loss of information. To address this issue, we introduced wavelet to replace the down-sampling operation thus reduce the image resolution. 

\begin{figure}
    \centering
    \includegraphics[width=8.5cm]{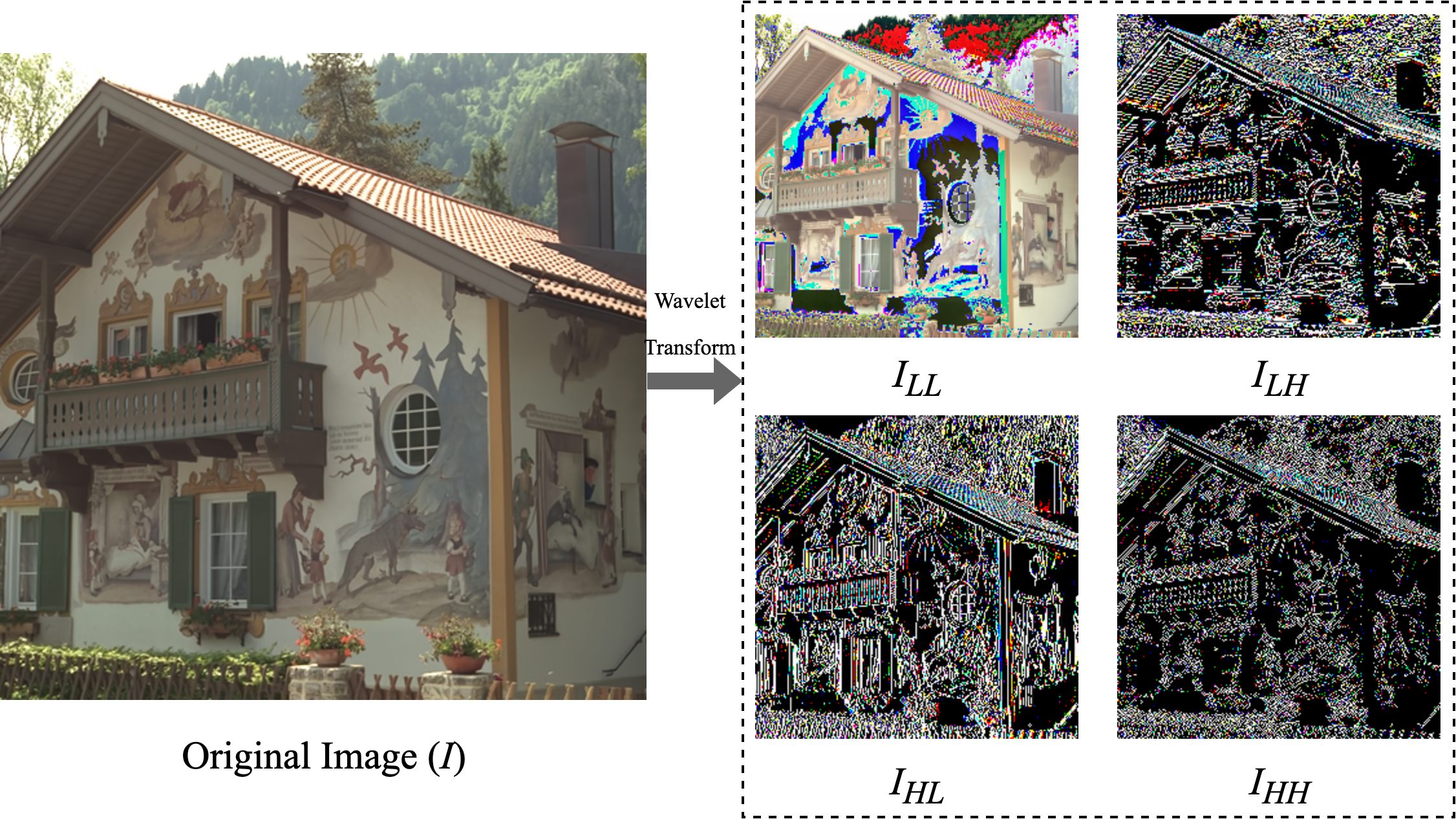}
    \caption{The schematic diagram of Discrete Wavelet Transform (DWT).}
    \label{wave}
\end{figure}

As shown in Fig.~\ref{wave}, when the Discrete Wavelet Transform (DWT) is applied to the image, the original image will be decomposed into four sub-images. Plenty of previous works have pointed out that these sub-bands have different frequencies, which mainly reflect the color of the filled area and the edge of the object. Specifically, $I_{LL}$ is the low-frequency information sub-band of the image, which is an approximation of the original image. $I_{LH}$ and $I_{HL}$ are the horizontal and vertical sub-bands of the image, reflecting the edge characteristics of these two directions. $I_{HH}$ is the diagonal sub-band of the image, reflecting the diagonal edge feature. Taking these sub-images as inputs of the model can guide the model to pay attention to frequency information and help to restore the texture details. Meanwhile, the connection between each sub-image can be established by the deep neural network, so that the model can extract deeper information. \textbf{Moreover, the wavelet is reversible and will not cause any loss of information, which is conducive to image restoration.} Therefore, we use Discrete Wavelet Transform (DWT) as the down-sampling module and use Inverse Wavelet Transform (IWT) as the up-sampling module in our EWT. In summary, the advantages of this method are: (1). The wavelet is reversible, so all information can be preserved through this sampling method; (2). Wavelet can capture the frequency and position information of the image, which is beneficial to restore the detailed features of the image; (3). Using wavelet can reduce the image resolution thus reducing the GPU memory consumption. Meanwhile, this process will not produce redundant parameters and can speed up the inference speed of the model, which benefit for efficient model building. (4). Wavelet will relatively increase the receptive area of the receptive field, so that the model can obtain richer features, which is benefit for image restoration.

\subsection{Multi-level Feature Aggregation Module}
As the core component of the entire model, Multi-level Feature Aggregation Module (MFAM) is specially designed for feature extraction and aggregation in the wavelet domain. As shown in Fig.~\ref{ewt}, MFAM consists of a series of DFEBs and a ConvBlock, which are responsible for the extraction and aggregation of features at different levels of the image, respectively. Different from the current methods simply stacking Transformer layers, we carefully design a double-branched structural unit (DFEB), and adopt the dense connection to combine the outputs of each DFEB. In this way, the hierarchical features of the model can be better aggregated to enhance the feature representation. Then, a ConvBlock is applied to incorporate these features:
\begin{equation}
    F_{d}=\sum_{i=1}^{N}F_{i},
\end{equation}
\begin{equation}
    F^{'}_{d}=f_{ConvB}(F_{d}),
\end{equation}
where $F_{i}$ represents the output of the $i$-th DFEB, $f_{ConvB}$ denotes the ConvBlock, and $F^{'}_{d}$ denotes the aggregated features. Finally, the global residual learning strategy is applied
\begin{equation}
    F_{out}=F_{in} + F^{'}_{d}.
\end{equation}

\textbf{Dual-stream Feature Extraction Block (DFEB)}: Most Transformer-based methods limit the use of convolutional layers and only use it for feature aggregation or downsampling. However, we found that if the proportion of Transformer is too high, the model performance and resource consumption will be seriously unbalanced. This is because there are matrix operations on large tensors in Transformer, which will consume a huge of GPU computing and storage resources
\begin{equation}
    Attention(Q, K, V) = Softmax(Norm(QK^{T}))V.
\end{equation}

Our experiments also show that staking a large number of Transformers will not significantly improve the model performance. On the contrary, it will greatly increase the calculation time and GPU memory consumption of the model. Meanwhile, we find that the CNN-based method is significantly faster than the Transformer-based method. Moreover, as the most widely used neural network in computer vision, CNN has been well proven to have the natural ability to capture image information. In particular, CNNs can extract the positional information of images without the need for additional positional encoding embeddings while Transformer does not have the ability to encode location information. Although most visual Transformers have embedded the position-coding operation, most of these operations are designed by human intuition. Compared to the ability of CNN to automatically learn location information, this is far from enough. Therefore, directly replacing CNN with Transformer is a sub-optimal solution. In this work, we focus on elegantly combining CNN and Transformer to find a better solution.

\begin{figure}[t]
    \centering
    \includegraphics[width=8.8cm]{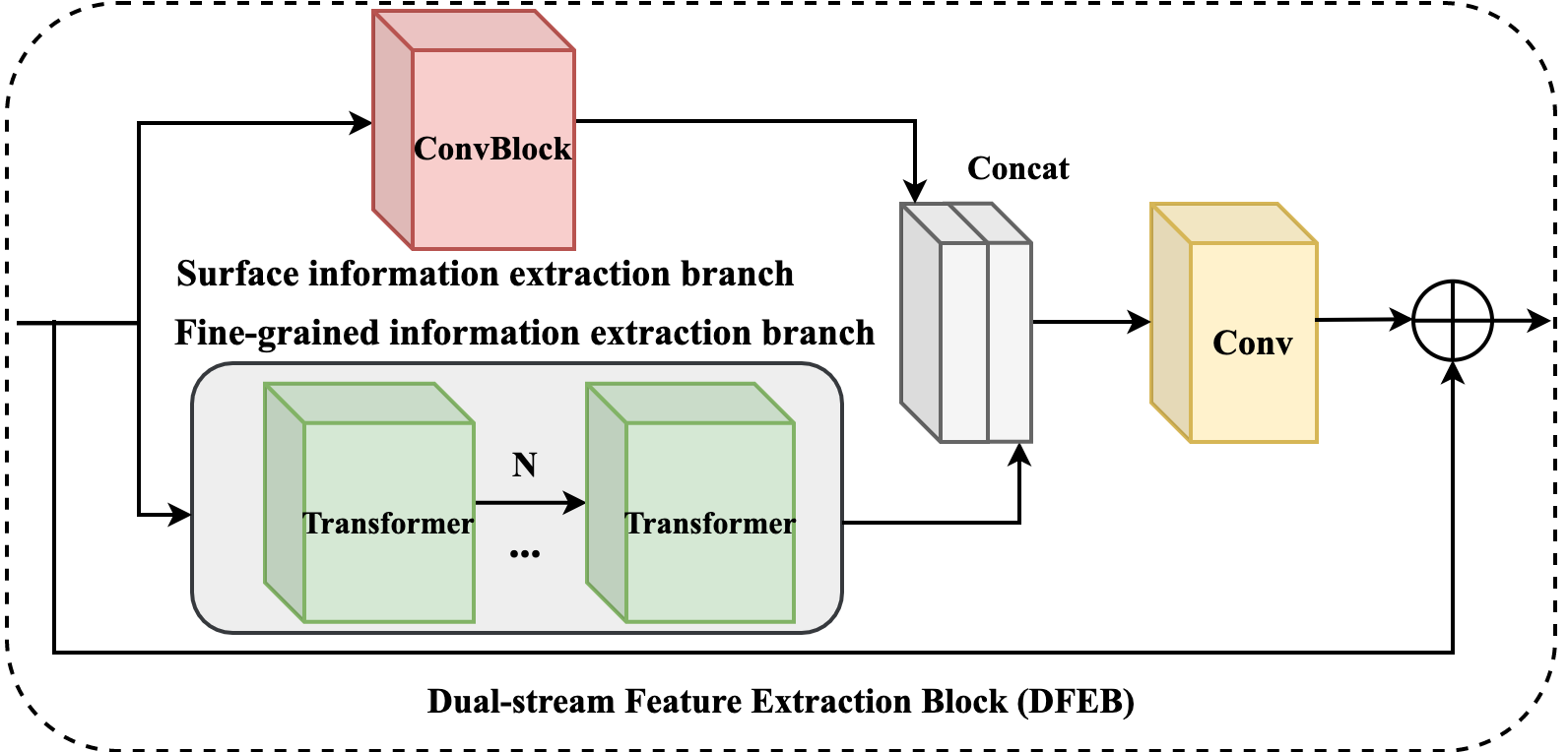}
    \caption{The complete architecture of Dual-stream Feature Extraction Block.}
    \label{DEFB}
\end{figure}

Inspired by the idea of multi-scale feature extraction, we find that the multi-branch structure can better guide the model to learn information at different scales. In addition, the parallelism of the multi-branch structure allows each branch to extract different features without interfering with each other, reducing the information dilution problem caused by excessive stacking of neural network modules. In a multi-scale CNN network, each branch is usually assigned a convolution kernel with different size to obtain different features under multiple receptive fields. In this work, we use Transformer as an alternative to multiple receptive fields. Specifically, we use Transformer and CNN as two branches to extract different features respectively, because CNN has strong local feature extraction ability and Transformer has better global encoding ability. Based on the above ideas, we designed a Dual-stream Feature Extraction Block (DFEB). DFEB is the most important component of MFAM, which is a dual-branch feature extraction module. The purpose of DFEB is to extract different levels of information and aggregate them to improve the expressive ability of the model.
\begin{figure}[!t]
    \centering
    \includegraphics[width=6.7cm]{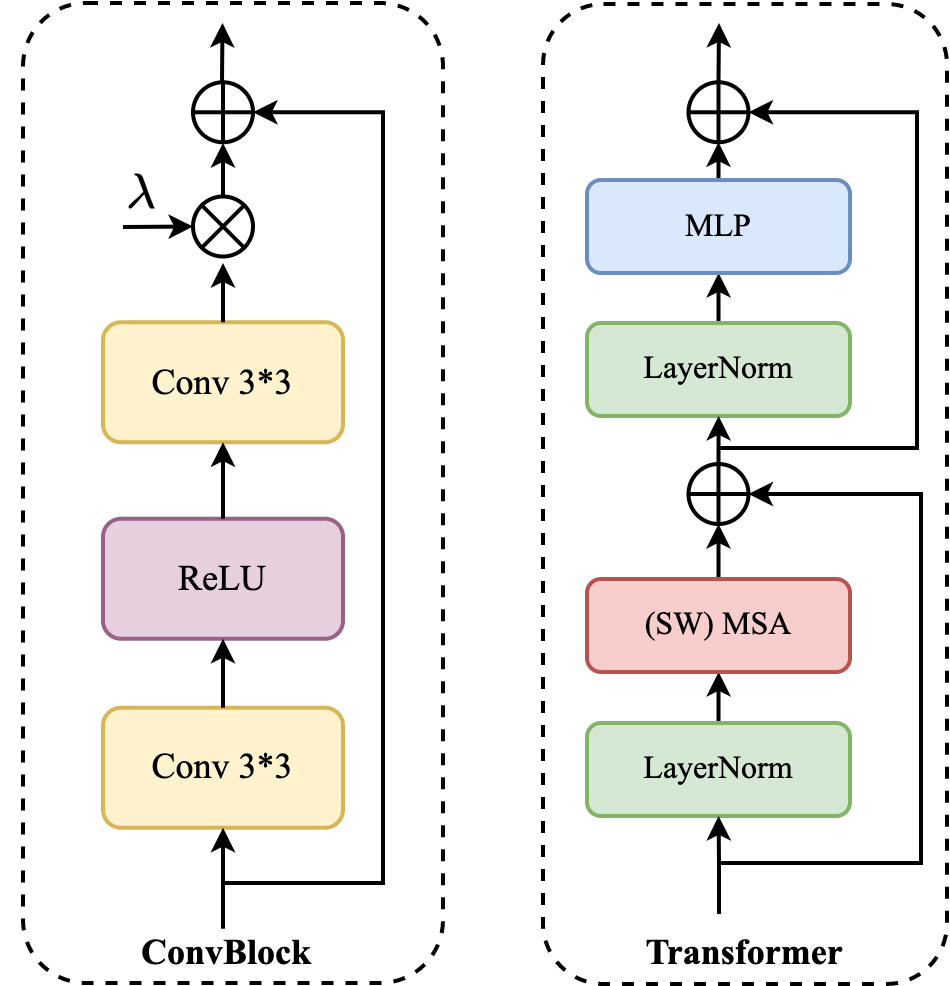}
    \caption{The complete architecture of ConvBlock and Transformer.}
    \label{Block}
\end{figure}
As shown in Fig.~\ref{DEFB}, DFEB contains two branches: surface information extraction branch and fine-gained information branch. When the features are sent to the DFEB, it will be divided into two groups, one group is used to extract rough features, and the other one is used to model the relationship among pixels and to learn the global information. Specifically, the surface information extraction branch only contains a ConvBlock (Fig.~\ref{Block}), which is a simple module composed of two convolutional layers and a ReLU activation function. This structure benefit for image restoration, especially for image surface information extraction. In the output, the weighted result of the convolution part is added to the input part to enhance the expression of shallow information. In the fine-gained information branch, we introduce the visual Transformer to extract the fine-grained information. Many methods have proved that Transformer can better model the pixel-level features of the image. However, since the image belongs to two-dimensional data, processing it in a serialized manner will destroy the location information of the image. Meanwhile, due to the huge overhead of the Transformer, it is unsuitable to directly model an entire feature map. Therefore, we borrowed the idea of Swin Transformer~\cite{r11} to decompose the feature map into smaller windows. Meanwhile, the window displacement mechanism is also be applied to enhance the information flow and interaction between windows. As shown in Fig.~\ref{Block}, (SW) MAS denotes the (Shift Window) Multi-Head Self-Attention mechanism proposed by Swin Transformer. Considering that the working mechanism of CNN and Transformer is different, adding the output features of these two branches directly will lead to information confusion. Therefore, we concatenate the output of CNN and Transformer to get rich features, and then a convolutional layer is used to weight and fuse different features to guide the module to learn useful features adaptively.

\begin{figure*}[!h]
\centering
\begin{minipage}[c]{0.16\textwidth}
\includegraphics[width=2.98cm,height=2cm]{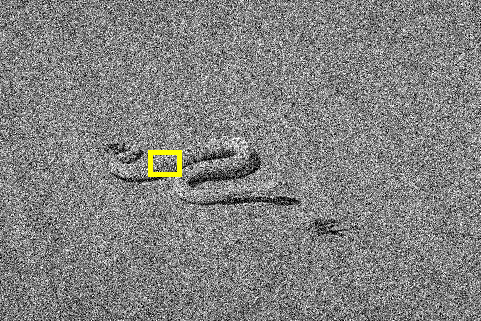}
\centerline{test036}
\end{minipage}
\begin{minipage}[c]{0.16\textwidth}
\includegraphics[width=2.98cm,height=2cm]{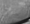}
\centerline{DnCNN: 26.30dB}
\end{minipage}
\begin{minipage}[c]{0.16\textwidth}
\includegraphics[width=2.98cm,height=2cm]{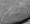}
\centerline{FFDNet: 26.27dB}
\end{minipage}
\begin{minipage}[c]{0.16\textwidth}
\includegraphics[width=2.98cm,height=2cm]{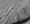}
\centerline{MLEFGN: 26.35dB}
\end{minipage}
\begin{minipage}[c]{0.16\textwidth}
\includegraphics[width=2.98cm,height=2cm]{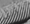}
\centerline{\textbf{EWT: 26.45dB}}
\end{minipage}
\begin{minipage}[c]{0.16\textwidth}
\includegraphics[width=2.98cm,height=2cm]{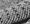}
\centerline{GT: PSNR}
\end{minipage}

\begin{minipage}[c]{1\textwidth}
\end{minipage}

\begin{minipage}[c]{0.16\textwidth}
\includegraphics[width=2.98cm,height=2cm]{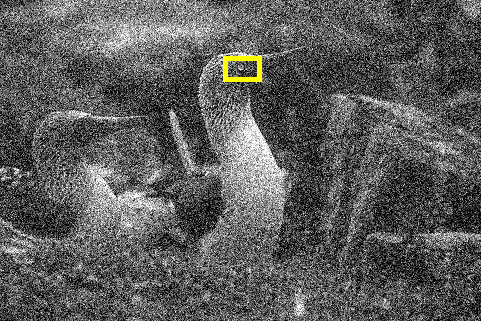}
\centerline{test004}
\end{minipage}
\begin{minipage}[c]{0.16\textwidth}
\includegraphics[width=2.98cm,height=2cm]{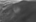}
\centerline{DnCNN: 27.60dB}
\end{minipage}
\begin{minipage}[c]{0.16\textwidth}
\includegraphics[width=2.98cm,height=2cm]{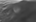}
\centerline{FFDNet: 27.71dB}
\end{minipage}
\begin{minipage}[c]{0.16\textwidth}
\includegraphics[width=2.98cm,height=2cm]{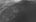}
\centerline{MLEFGN: 27.85dB}
\end{minipage}
\begin{minipage}[c]{0.16\textwidth}
\includegraphics[width=2.98cm,height=2cm]{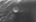}
\centerline{\textbf{EWT: 27.94dB}}
\end{minipage}
\begin{minipage}[c]{0.16\textwidth}
\includegraphics[width=2.98cm,height=2cm]{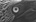}
\centerline{GT: PSNR}
\end{minipage}

\begin{minipage}[c]{1\textwidth}
\end{minipage}

\begin{minipage}[c]{0.16\textwidth}
\includegraphics[width=2.98cm,height=2cm]{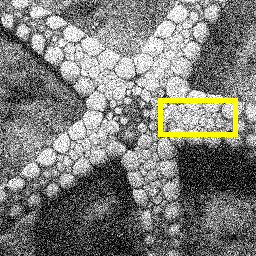}
\centerline{set12-04}
\end{minipage}
\begin{minipage}[c]{0.16\textwidth}
\includegraphics[width=2.98cm,height=2cm]{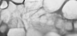}
\centerline{DnCNN: 25.70dB}
\end{minipage}
\begin{minipage}[c]{0.16\textwidth}
\includegraphics[width=2.98cm,height=2cm]{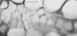}
\centerline{FFDNet: 25.77dB}
\end{minipage}
\begin{minipage}[c]{0.16\textwidth}
\includegraphics[width=2.98cm,height=2cm]{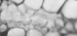}
\centerline{MLEFGN: 25.77dB}
\end{minipage}
\begin{minipage}[c]{0.16\textwidth}
\includegraphics[width=2.98cm,height=2cm]{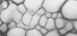}
\centerline{\textbf{EWT: 26.32dB}}
\end{minipage}
\begin{minipage}[c]{0.16\textwidth}
\includegraphics[width=2.98cm,height=2cm]{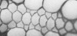}
\centerline{GT: PSNR}
\end{minipage}

\caption{Visual comparison on grayscale images with $\sigma=50$. Obviously, our EWT can reconstruct high-quality noise-free images with clear edges.}
\label{visual-grayscale}
\end{figure*}

\section{Experiments}\label{EXP}
\subsection{Datasets}
In this paper, we use 800 training images in DIV2K~\cite{DIV2K} as the training set. For evaluation, we choose six benchmark test sets, including Set12~\cite{zeyde2010single}, BSD68~\cite{roth2005fields}, Kodak24~\cite{franzen1999kodak}, CBSD68~\cite{martin2001database}, and Urban100~\cite{huang2015single}. In addition, we choose additive white Gaussian noise (AWGN) as our research object since AWGN is the best approximation of the real mixture noise, which can simulate the disturbance of real noise to the image. Following previous works, we use Set12, BSD68, and Urban100 to evaluate the performance of EWT in grayscale images, and use Kodak24, CBSD68, and Urban100 to evaluate the denoising effect of model on color images. Meanwhile, to further verify the effectiveness and robustness of EWT, we utilize SIDD~\cite{abdelhamed2018high} and RNI15~\cite{lebrun2015noise} to evaluate the denoising performance of the model in the real image denoising task.

\subsection{Implementation Details}
Before training, we generate noisy images by adding AWGN with different noise levels. To verify the effectiveness of the model, we set the noise level $\sigma$ = 15, 25, and 50 for grayscale images and set $\sigma$ = 10, 30, and 50 for color images. During training, we randomly choose 16 noisy patches as inputs and these patches are randomly rotated and flipped to enhance the data. In addition, EWT is implemented with PyTorch framework and updated with the Adam optimizer.

In the final model, we use a single-scale wavelet to sample the image. The size of all convolution kernels in the model is $3 \times 3$, the $\lambda$ in the ConvBlock is set to $0.1$, and the embedding dimension of MFAM is set to $180$. In addition, we use $4$ DFEB in MFAM, and each DFEB contains $1$ ConvBlock and $6$ Transformer blocks. In the Transformer, the window size is $8$, the number of attention heads is $6$, and the MLP dimension is as twice as the embedding dimension.
\begin{table*}[!t]
\small
\centering
\setlength{\tabcolsep}{2.2mm}
\caption{PSNR (dB) comparison with other classic SID methods on grayscale image test datasets. The best results are \textbf{highlighted}.}
\begin{tabular}{|l|ccc|ccc|ccc|c|}
\hline
Method                 & \multicolumn{3}{c|}{Set12}                                                                             & \multicolumn{3}{c|}{BSD68}                                                           & \multicolumn{3}{c|}{Urban100}                                                  &   \multirow{2}{*}{Average}  \\ \cline{1-10}
Noise Level & \multicolumn{1}{c|}{$\sigma$ = 15}             & \multicolumn{1}{c|}{$\sigma$ = 25}             & $\sigma$ = 50                         & \multicolumn{1}{c|}{$\sigma$ = 15}    & \multicolumn{1}{c|}{$\sigma$ = 25}    & $\sigma$ = 50                         & \multicolumn{1}{c|}{$\sigma$ = 15}    & \multicolumn{1}{c|}{$\sigma$ = 25}    & $\sigma$ = 50                     &    \\ \hline
BM3D~\cite{dabov2007image}                   & \multicolumn{1}{c|}{32.37dB}          & \multicolumn{1}{c|}{29.97dB}          & 26.72dB                      & \multicolumn{1}{c|}{31.08dB} & \multicolumn{1}{c|}{28.57dB} & 25.60dB                      & \multicolumn{1}{c|}{32.35dB} & \multicolumn{1}{c|}{29.70dB} & 25.95dB         &   29.15dB      \\ \hline
WNNM~\cite{gu2014weighted}                   & \multicolumn{1}{c|}{32.70dB}          & \multicolumn{1}{c|}{30.28dB}          & 27.05dB                      & \multicolumn{1}{c|}{31.37dB} & \multicolumn{1}{c|}{28.83dB} & 25.87dB                      & \multicolumn{1}{c|}{32.97dB} & \multicolumn{1}{c|}{30.39dB} & 26.83dB              &   29.59dB     \\ \hline
IRCNN~\cite{zhang2017learning}                  & \multicolumn{1}{c|}{32.76dB}          & \multicolumn{1}{c|}{30.37dB}          & 27.12dB                      & \multicolumn{1}{c|}{31.63dB} & \multicolumn{1}{c|}{29.15dB} & 26.19dB                      & \multicolumn{1}{c|}{32.46dB} & \multicolumn{1}{c|}{29.80dB} & 26.22dB               &   29.52dB    \\ \hline
DnCNN~\cite{r13}                  & \multicolumn{1}{c|}{32.86dB}          & \multicolumn{1}{c|}{30.44dB}          & 27.18dB                      & \multicolumn{1}{c|}{31.73dB} & \multicolumn{1}{c|}{29.23dB} & 26.23dB                      & \multicolumn{1}{c|}{32.64dB} & \multicolumn{1}{c|}{29.95dB} & 26.26dB                &   29.61dB   \\ \hline
FFDNet~\cite{zhang2018ffdnet}                 & \multicolumn{1}{c|}{32.75dB}          & \multicolumn{1}{c|}{30.43dB}          & 27.32dB                      & \multicolumn{1}{c|}{31.63dB} & \multicolumn{1}{c|}{29.19dB} & 26.29dB                      & \multicolumn{1}{c|}{32.40dB} & \multicolumn{1}{c|}{29.90dB} & 26.50dB                &   29.60dB   \\ \hline
RED30~\cite{mao2016image}                  & \multicolumn{1}{c|}{32.83dB}          & \multicolumn{1}{c|}{30.48dB}          & \multicolumn{1}{c|}{27.34dB} & \multicolumn{1}{c|}{31.72dB} & \multicolumn{1}{c|}{29.26dB} & \multicolumn{1}{c|}{26.35dB} & \multicolumn{1}{c|}{32.75dB} & \multicolumn{1}{c|}{30.21dB} & \multicolumn{1}{c|}{26.48dB}  &  29.71dB \\ \hline
MLEFGN~\cite{fang2020multilevel}                 & \multicolumn{1}{c|}{33.04dB}          & \multicolumn{1}{c|}{30.66dB}          & 27.54dB                      & \multicolumn{1}{c|}{31.81dB} & \multicolumn{1}{c|}{29.34dB} & 26.39dB                      & \multicolumn{1}{c|}{33.21dB} & \multicolumn{1}{c|}{30.64dB} & 27.22dB                  &  29.98dB  \\ \hline
MWCNN~\cite{r12}                  & \multicolumn{1}{c|}{33.15dB}          & \multicolumn{1}{c|}{30.79dB}          & 27.74dB                      & \multicolumn{1}{c|}{31.86dB} & \multicolumn{1}{c|}{\textbf{29.41dB}} & \textbf{26.53dB}                      & \multicolumn{1}{c|}{33.17dB} & \multicolumn{1}{c|}{30.66dB} & 27.42dB                 &  30.08dB   \\ \hline
EWT (Ours)             & \multicolumn{1}{c|}{\textbf{33.23dB}} & \multicolumn{1}{c|}{\textbf{30.86dB}} & \textbf{27.80dB}                      & \multicolumn{1}{c|}{\textbf{31.87dB}} & \multicolumn{1}{c|}{29.40dB} & 26.47dB                      & \multicolumn{1}{c|}{\textbf{33.54dB}} & \multicolumn{1}{c|}{\textbf{31.08dB}} & \textbf{27.70dB}                  &   \textbf{30.22dB}  \\ \hline
\end{tabular}
\label{Grayimage}
\end{table*}

\begin{table*}[!t]
\small
\centering
\setlength{\tabcolsep}{2.2mm}
\caption{PSNR (dB) comparison with other classic SID methods on color image test datasets. The best results are \textbf{highlighted}.}
\begin{tabular}{|l|ccc|ccc|ccc|c|}
\hline
Method     & \multicolumn{3}{c|}{Kodak24}                                                                           & \multicolumn{3}{c|}{CBSD68}                                                                            & \multicolumn{3}{c|}{CUrban100}                                                                      &   \multirow{2}{*}{Average}  \\ \cline{1-10}
Noise Level           & \multicolumn{1}{c|}{$\sigma$ = 10}             & \multicolumn{1}{c|}{$\sigma$ = 30}             & $\sigma$ = 50                         & \multicolumn{1}{c|}{$\sigma$ = 10}             & \multicolumn{1}{c|}{$\sigma$ = 30}             & $\sigma$ = 50                         & \multicolumn{1}{c|}{$\sigma$ = 10}             & \multicolumn{1}{c|}{$\sigma$ = 30}             & $\sigma$ = 50           &              \\ \hline
CBM3D~\cite{dabov2007image}      & \multicolumn{1}{c|}{36.57dB}          & \multicolumn{1}{c|}{30.89dB}          & 28.63dB                      & \multicolumn{1}{c|}{35.91dB}          & \multicolumn{1}{c|}{29.73dB}          & 27.38dB                      & \multicolumn{1}{c|}{36.00dB}          & \multicolumn{1}{c|}{30.36dB}          & 27.94dB               &   31.49dB    \\ \hline
IRCNN~\cite{zhang2017learning}      & \multicolumn{1}{c|}{36.70dB}          & \multicolumn{1}{c|}{31.24dB}          & 28.93dB                      & \multicolumn{1}{c|}{36.06dB}          & \multicolumn{1}{c|}{30.22dB}          & 27.86dB                      & \multicolumn{1}{c|}{35.81dB}          & \multicolumn{1}{c|}{30.28dB}          & 27.69dB            &    31.64dB      \\ \hline
DnCNN~\cite{r13}      & \multicolumn{1}{c|}{36.98dB}          & \multicolumn{1}{c|}{31.39dB}          & 29.16dB                      & \multicolumn{1}{c|}{36.31dB}          & \multicolumn{1}{c|}{30.40dB}          & 28.01dB                      & \multicolumn{1}{c|}{36.21dB}          & \multicolumn{1}{c|}{30.28dB}          & 28.16dB                 &   31.87dB  \\ \hline

FFDNet~\cite{zhang2018ffdnet}     & \multicolumn{1}{c|}{36.81dB}          & \multicolumn{1}{c|}{31.39dB}          & 29.10dB                      & \multicolumn{1}{c|}{36.14dB}          & \multicolumn{1}{c|}{30.31dB}          & 27.96dB                      & \multicolumn{1}{c|}{35.77dB}          & \multicolumn{1}{c|}{30.53dB}          & 28.05dB             &   31.78dB      \\ \hline
MLEFGN~\cite{fang2020multilevel}       & \multicolumn{1}{c|}{37.04dB}          & \multicolumn{1}{c|}{31.67dB}          & 29.38dB                      & \multicolumn{1}{c|}{36.37dB}          & \multicolumn{1}{c|}{30.56dB}          & 28.21dB                      & \multicolumn{1}{c|}{36.42dB}          & \multicolumn{1}{c|}{31.32dB}          & 28.92dB               &   32.21dB    \\ \hline
RNAN~\cite{zhang2019residual}       & \multicolumn{1}{c|}{37.24dB}          & \multicolumn{1}{c|}{31.86dB}          & \multicolumn{1}{c|}{29.58dB} & \multicolumn{1}{c|}{36.43dB}          & \multicolumn{1}{c|}{30.63dB}          & \multicolumn{1}{c|}{28.27dB} & \multicolumn{1}{c|}{36.59dB}          & \multicolumn{1}{c|}{31.50dB}          & \multicolumn{1}{c|}{29.08dB}  &  32.35dB \\ \hline
RDN~\cite{zhang2020residual}        & \multicolumn{1}{c|}{\textbf{37.31dB}} & \multicolumn{1}{c|}{\textbf{31.94dB}} & \textbf{29.66dB}             & \multicolumn{1}{c|}{36.47dB}          & \multicolumn{1}{c|}{30.67dB}          & 28.31dB                      & \multicolumn{1}{c|}{\textbf{36.69dB}}          & \multicolumn{1}{c|}{31.69dB}          & 29.29dB                &    32.45dB  \\ \hline
EWT (Ours) & \multicolumn{1}{c|}{37.24dB} & \multicolumn{1}{c|}{31.90dB} & 29.63dB                      & \multicolumn{1}{c|}{\textbf{36.49dB}} & \multicolumn{1}{c|}{\textbf{30.71dB}} & \textbf{28.36dB}             & \multicolumn{1}{c|}{\textbf{36.69dB}} & \multicolumn{1}{c|}{\textbf{31.81dB}} & \textbf{29.52dB}       &    \textbf{32.48dB}  \\ \hline
\end{tabular}
\label{Colorimage}
\end{table*}

\begin{figure*}[!h]
\centering
\begin{minipage}[c]{0.16\textwidth}
\includegraphics[width=2.98cm,height=2.13cm]{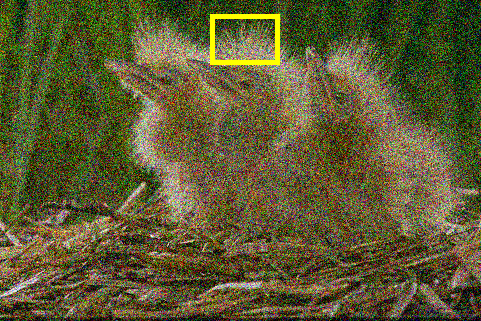}
\centerline{163085}
\end{minipage}
\begin{minipage}[c]{0.16\textwidth}
\includegraphics[width=2.98cm]{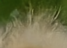}
\centerline{DnCNN: 28.35dB}
\end{minipage}
\begin{minipage}[c]{0.16\textwidth}
\includegraphics[width=2.98cm]{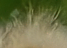}
\centerline{ADNet: 28.35dB}
\end{minipage}
\begin{minipage}[c]{0.16\textwidth}
\includegraphics[width=2.98cm]{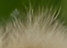}
\centerline{MLEFGN: 28.96dB}
\end{minipage}
\begin{minipage}[c]{0.16\textwidth}
\includegraphics[width=2.98cm]{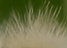}
\centerline{\textbf{EWT: 29.09dB}}
\end{minipage}
\begin{minipage}[c]{0.16\textwidth}
\includegraphics[width=2.98cm]{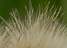}
\centerline{GT: PSNR}
\end{minipage}

\begin{minipage}[c]{1\textwidth}
\end{minipage}

\begin{minipage}[c]{0.16\textwidth}
\includegraphics[width=2.98cm,height=1.7cm]{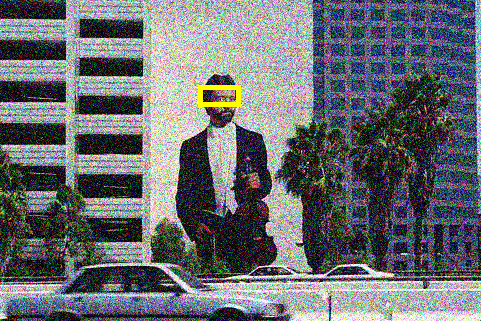}
\centerline{119082}
\end{minipage}
\begin{minipage}[c]{0.16\textwidth}
\includegraphics[width=2.98cm,height=1.7cm]{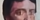}
\centerline{DnCNN: 27.27dB}
\end{minipage}
\begin{minipage}[c]{0.16\textwidth}
\includegraphics[width=2.98cm,height=1.7cm]{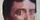}
\centerline{ADNet: 27.62dB}
\end{minipage}
\begin{minipage}[c]{0.16\textwidth}
\includegraphics[width=2.98cm,height=1.7cm]{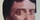}
\centerline{MLEFGN: 28.21dB}
\end{minipage}
\begin{minipage}[c]{0.16\textwidth}
\includegraphics[width=2.98cm,height=1.7cm]{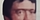}
\centerline{\textbf{EWT: 28.46dB}}
\end{minipage}
\begin{minipage}[c]{0.16\textwidth}
\includegraphics[width=2.98cm,height=1.7cm]{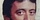}
\centerline{GT: PSNR}
\end{minipage}

\begin{minipage}[c]{1\textwidth}
\end{minipage}

\begin{minipage}[c]{0.16\textwidth}
\includegraphics[width=3.0cm,height=3.62cm]{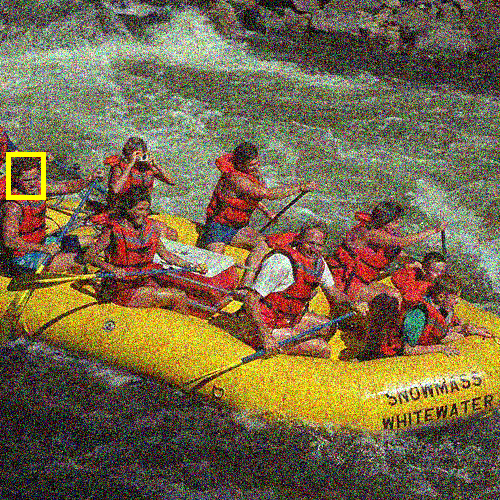}
\centerline{kodim14}
\end{minipage}
\begin{minipage}[c]{0.16\textwidth}
\includegraphics[width=3.0cm]{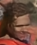}
\centerline{DnCNN: 26.78dB}
\end{minipage}
\begin{minipage}[c]{0.16\textwidth}
\includegraphics[width=3.0cm]{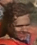}
\centerline{ADNet: 26.79dB}
\end{minipage}
\begin{minipage}[c]{0.16\textwidth}
\includegraphics[width=3.0cm]{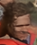}
\centerline{MLEFGN: 27.27dB}
\end{minipage}
\begin{minipage}[c]{0.16\textwidth}
\includegraphics[width=3.0cm]{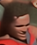}
\centerline{\textbf{EWT: 27.26dB}}
\end{minipage}
\begin{minipage}[c]{0.16\textwidth}
\includegraphics[width=3.0cm]{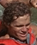}
\centerline{GT: PSNR}
\end{minipage}
\caption{Visual comparison on color images with $\sigma=50$. Obviously, our EWT can reconstruct high-quality noise-free images with clear edges.}
\label{visual-color}
\end{figure*}

\subsection{Comparisons with State-of-the-art Methods}
\quad \textbf{Gray-scale Image Denoising:} In Tabel~\ref{Grayimage}, we report the PSNR results of different SID methods on three benchmark test sets. Obviously, EWT achieves competitive results and the best average results on these test sets with different noise levels. It is worth noting that MWCNN is also a wavelet-based SID model, which achieved slightly better results than EWT on BSD68 ($\sigma$ = 25 and 50). However, it cannot be ignored that the results of MWCNN under other test sets are all worse than our EWT, and the average result is 0.14dB worse than EWT. Meanwhile, MWCNN uses multiple training sets to train the model, which contains 5744 images (7 times of our training images). Under this disparity, EWT still achieves close or better results, which fully demonstrates its effectiveness.

In Fig.~\ref{visual-grayscale}, we provide the visual comparison of the denoised images with noise levels $\sigma$ = 50. In this part, we choose three most representative CNN-based image denoising methods for comparison, including DnCNN~\cite{r13}, FFDNet~\cite{zhang2018ffdnet}, and MLEFGN~\cite{fang2020multilevel}. Among them, DnCNN and FFDNet are the two most classic CNN-based image denoising models. According to the figure, we can clearly observe that the images reconstructed by DnCNN and FDDNet are too smooth, and they have lost texture details and edge information. As for MLEFGN, it can reconstruct 
more clear noise-free images, but the edges of the image are not accurate and complete enough. In contrast, our EWT can reconstruct high-quality images with clear and accurate texture details and edges. This further illustrates the effectiveness and excellence of EWT.
\begin{table*}[!h]
\small
\centering
\setlength{\tabcolsep}{6.45mm}
\caption{Quantitative comparison with other Transformer methods on Poisson noise and Speckle noise.}
\begin{tabular}{|c|ccc|ccc|}
\hline
Noise Level & \multicolumn{3}{c|}{Poisson}                                          & \multicolumn{3}{c|}{Speckle}                                         \\ \hline
         Method    & \multicolumn{1}{l|}{Kodak24} & \multicolumn{1}{l|}{CBSD68} & Urban100 & \multicolumn{1}{l|}{Kodak24} & \multicolumn{1}{l|}{CBSD68} & Urban100 \\ \hline
SwinIR~\cite{r19}     & \multicolumn{1}{l|}{37.09dB}  & \multicolumn{1}{l|}{36.44dB} & 36.58dB   & \multicolumn{1}{l|}{31.07dB}  & \multicolumn{1}{l|}{29.87dB} & 29.94dB   \\ \hline
Uformer~\cite{wang2022uformer}     & \multicolumn{1}{l|}{36.80dB}  & \multicolumn{1}{l|}{36.08dB} & 36.20dB   & \multicolumn{1}{l|}{30.71dB}  & \multicolumn{1}{l|}{29.42dB} & 29.72dB   \\ \hline
Restormer~\cite{zamir2022restormer}    & \multicolumn{1}{l|}{37.14dB}  & \multicolumn{1}{l|}{36.52dB} & 36.66dB   & \multicolumn{1}{l|}{31.01dB}  & \multicolumn{1}{l|}{29.85dB} & 29.90dB   \\ \hline
EWT          & \multicolumn{1}{l|}{\textbf{37.20dB}}  & \multicolumn{1}{l|}{\textbf{36.52dB}} & \textbf{36.61dB}   & \multicolumn{1}{l|}{\textbf{31.24dB}}  & \multicolumn{1}{l|}{\textbf{29.98dB}} & \textbf{29.90dB}   \\ \hline
\end{tabular}
\label{possion}
\end{table*}
\textbf{Color Image Denoising:} As for color image denoising, we use Kodak24, CBSD68, and CUrban100 to verify its performance. In this part, we choose three most representative CNN-based image denoising methods for comparison, including DnCNN~\cite{r13}, ADNet~\cite{zhang2018ffdnet}, and MLEFGN~\cite{fang2020multilevel}. According to TABLE~\ref{Colorimage}, we can clearly observe that our EWT still achieves excellent results on color images, especially on Urban100. Among them, RDN is recognized as one of the most advanced SID models, which is specially designed for color image denoising. Compared with it, our EWT achieved close results on Kodak24 and better results on CBSD68 and CUrban100. It is worth noting that our EWT achieves better average result than RDN with only half of the parameters (EWT: 11M vs RDN: 22M). These results fully demonstrate the denoising ability of EWT on color images, further validating the effectiveness of EWT.

In Fig.~\ref{visual-color}, we provide the visual comparisons of the denoised images with $\sigma$ = 50 on CBSD68. In this part, we also choose three most representative CNN-based image denoising methods for comparison, including DnCNN~\cite{r13}, ADNet~\cite{tian2020attention}, and MLEFGN~\cite{fang2020multilevel}. Obviously, our EWT can reconstruct high-quality noise-free images with sharper and more accurate edges. Taking the human face as an example, our EWT can reconstruct clearer and more accurate contours. This is due to the fact that the Transformer introduced in EWT can capture the global information of the face, thereby reconstructing high-quality face. All these results further illustrate the effectiveness of the proposed EWT.

\begin{figure*}[!t]
\centering
\begin{minipage}[c]{0.195\textwidth}
\includegraphics[width=3.6cm,height=1.9cm]{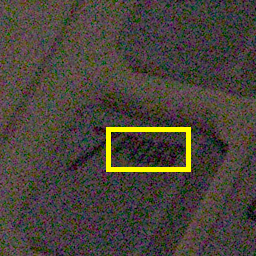}
\centerline{0012-0010}
\centerline{}
\end{minipage}
\begin{minipage}[c]{0.195\textwidth}
\includegraphics[width=3.6cm,height=1.9cm]{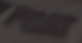}
\centerline{RIDNet: 38.41dB}
\centerline{}
\end{minipage}
\begin{minipage}[c]{0.195\textwidth}
\includegraphics[width=3.6cm,height=1.9cm]{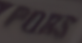}
\centerline{CycleISP: 39.77dB}
\centerline{}
\end{minipage}
\begin{minipage}[c]{0.195\textwidth}
\includegraphics[width=3.6cm,height=1.9cm]{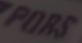}
\centerline{\textbf{EWT: 39.92dB}}
\centerline{}
\end{minipage}
\begin{minipage}[c]{0.195\textwidth}
\includegraphics[width=3.6cm,height=1.9cm]{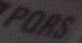}
\centerline{GT: PSNR}
\centerline{}

\end{minipage}
\begin{minipage}[c]{0.195\textwidth}
\includegraphics[width=3.6cm,height=1.9cm]{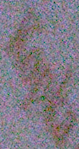}
\centerline{0033-0026}
\centerline{}
\end{minipage}
\begin{minipage}[c]{0.195\textwidth}
\includegraphics[width=3.6cm,height=1.9cm]{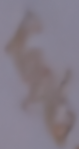}
\centerline{RIDNet: 38.78dB}
\centerline{}
\end{minipage}
\begin{minipage}[c]{0.195\textwidth}
\includegraphics[width=3.6cm,height=1.9cm]{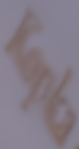}
\centerline{CycleISP: 40.21dB}
\centerline{}
\end{minipage}
\begin{minipage}[c]{0.195\textwidth}
\includegraphics[width=3.6cm,height=1.9cm]{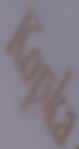}
\centerline{\textbf{EWT: 40.74dB}}
\centerline{}
\end{minipage}
\begin{minipage}[c]{0.195\textwidth}
\includegraphics[width=3.6cm,height=1.9cm]{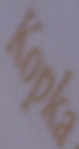}
\centerline{GT: PSNR}
\centerline{}
\end{minipage}
\caption{Visual comparison on real-noise images (SIDD~\cite{abdelhamed2018high}). Obviously, EWT can reconstruct high-quality noise-free images.}
\label{visual-real}
\end{figure*}

\begin{figure*}[!t]
\centering
\begin{minipage}[c]{0.195\textwidth}
\includegraphics[width=3.6cm,height=1.9cm]{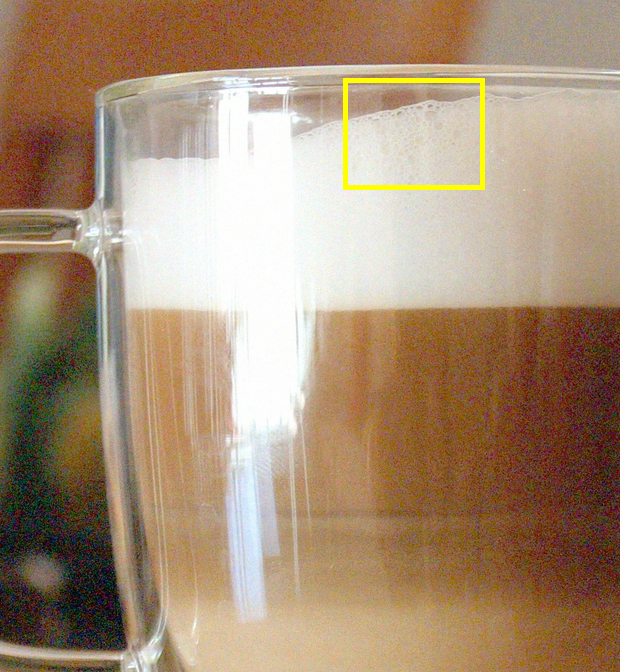}
\centerline{Glass}
\end{minipage}
\begin{minipage}[c]{0.195\textwidth}
\includegraphics[width=3.6cm,height=1.9cm]{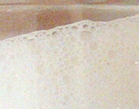}
\centerline{Noisy}
\end{minipage}
\begin{minipage}[c]{0.195\textwidth}
\includegraphics[width=3.6cm,height=1.9cm]{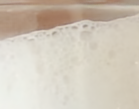}
\centerline{RIDNet}
\end{minipage}
\begin{minipage}[c]{0.195\textwidth}
\includegraphics[width=3.6cm,height=1.9cm]{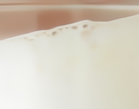}
\centerline{MLEFGN}
\end{minipage}
\begin{minipage}[c]{0.195\textwidth}
\includegraphics[width=3.6cm,height=1.9cm]{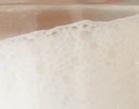}
\centerline{EWT (Ours)}
\end{minipage}
\caption{Visual comparison on real-noise images (RNI15~\cite{lebrun2015noise}). Obviously, EWT can reconstruct high-quality noise-free images.}
\label{RNI15}
\end{figure*}

\textbf{Restoration of Other Synthetic Noise: }The noise used in practical applications is usually more than Gaussian noise, and other noises are also very common, such as Poisson noise and Speckle noise. Since it has a more complex distribution, it also needs to be considered emphatically. In order to verify the general applicability of the method in this paper, TABLE ~\ref{possion} compares EWT with
three classic image restoration Transformer methods. The results show that EWT also performs well in other noisy images. This is due to the idea of combining wavelet transform in this paper, which ensures that the Transformer always maintains attention to the image texture details during the feature extraction process. This further validates the effectiveness of our proposed EWT, and also reflects the generality of EWT on different noisy images.

\begin{table*}[t]
\small
 \centering
\setlength{\tabcolsep}{1.4mm}
\caption{Real image denoising comparison with DnCNN~\cite{r13}, BM3D~\cite{dabov2007image}, CBDNet~\cite{guo2019toward}, RIDNet~\cite{anwar2019real}, AINDNet~\cite{kim2020transfer}, VDN~\cite{yue2019variational}, SADNet~\cite{chang2020spatial}, DANet+~\cite{yue2020dual}, CycleISR~\cite{zamir2020cycleisp}, DeamNet~\cite{ren2021adaptive} on SIDD~\cite{abdelhamed2018high} ($\star $ denote the model using additional training sets to train the model).}
  \begin{tabular}{@{}c|ccccccccccc@{}}
   \toprule
   Method        &DnCNN    &BM3D   &CBDNet$\star $  &RIDNet$\star $  &AINDNet$\star $    &VDN       &SADNet$\star $     &DANet+$\star $ & CycleISR$\star $   &DeamNet$\star $      &EWT (Ours)                \\ 
   \midrule
   PSNR       &23.66dB      &25.65dB   &30.78dB  & 38.71dB  &38.95dB   &39.28dB      &39.46dB          &39.47dB    &39.52dB  &39.35dB    &\textbf{39.52dB}         \\
   \bottomrule 
  \end{tabular}
  \label{real}
\end{table*}

\textbf{Real Image Denoising: }Real image denoising is a more difficult task since real image noise comes from multiple sources. In this part, real noisy images are used to further assess the practicability of the proposed EWT. In TABLE~\ref{real}, we provide PSNR comparisons of EWT with other models specially designed for real image denoising. Among them, * denote the model using additional training sets to train the model. Obviously, our model still achieves the best results even without using additional training sets. This further validates the effectiveness and versatility of our EWT. In addition, we also provide the visual comparison on SIDD~\cite{abdelhamed2018high} and RNI15~\cite{lebrun2015noise} sets in Figs.~\ref{visual-real} and~\ref{RNI15}, respectively. Obviously, our EWT still can reconstruct high-quality noise-free images. This shows that EWT also performs well on the real image denoising task.

\begin{table}[!h]
\small
	\centering
	\setlength{\tabcolsep}{1.4mm}
    \caption{Comparison with SwinIR* on Kodak24 ($\sigma$ = 30, color).}
		\begin{tabular}{@{}cccccc@{}}
			\toprule
			Method   & Patchsize	    &GPU       &   Time	         &Params     &PSNR	      	    \\ 
			\midrule
			SwinIR*    & 56  	&18432MiB            &53.29s                  &5.17M      & \textbf{31.79dB}              \\
   DWT+SwinIR*    & 56  	&8276MiB            &12.52s                  &5.20M      & \textbf{31.57dB}              \\
			EWT*        & 56   	&\textbf{6347MiB}     &\textbf{9.14s}         &5.18M      &31.73dB     \\ 
			\bottomrule 
		\end{tabular}
		\label{Compare with SwinIR}
\end{table}

\section{Ablation Studies}\label{AS}
\subsection{Wavelet Investigations} 
In our method, the wavelet plays a vital role in shortening the execution time and GPU memory consumption. To verify this statement, we compare it with SwinIR~\cite{r19}. SwinIR is a famous Transformer-based image restoration model, which does not use wavelet or other operations to change the image resolution. It is worth noting that SwinIR uses additional training sets and the GPU memory required for it exceeds the maximum limit of our device. For a fair comparison, the embedding dimension of MFAM in SinwIR and EWT are both reduced from 180 to 120, and these two models are retrained under the same data set and settings. In addition, we also consider the combination of DWT and SwinIR to further illustrate the effectiveness of EWT and the rationality of its structure. Meanwhile, we label these two modified models as SwinIR* and EWT*, respectively. According to TABLE~\ref{Compare with SwinIR}, we can clearly observe that EWT* and SwinIR* have a similar number of parameters, and EWT achieves close PSNR results to SwinIR with only 1/6 running time and 1/3 GPU memory. In addition, we also noticed that directly combine DWT with SwinIR will degrade the performance of the model since it does not optimize the structural design of the network. Contrastly, our EWT achieves better results due to its well-designed network
structure and effective DFEB. This huge breakthrough fully demonstrated the advantages of wavelet and further verified the advancement and effectiveness of EWT.

\begin{table}[t]
\centering
\small
\setlength{\tabcolsep}{0.8mm}
\caption{Study of Multi-level Wavelet (Kodak24, $\sigma$ = 30, color).}
\begin{tabular}{@{}c|ccccccc@{}}
\toprule
Case &Multi-Level   &Time   &Patchsize &GPU  &FLOPs  &PSNR           \\ 
\midrule
1    &$\times1$     &11.96s &64        &7636MiB     &17.82G &\textbf{31.78dB} \\
2    &$\times2$     &3.09s  &64        &3658MiB      &4.50G  &31.62dB          \\
3    &$\times3$     &1.91s  &64        &2758MiB      &1.18G  &27.94dB          \\ 
\bottomrule 
\end{tabular}
\label{Wavelet}
\end{table}

In order to further verify the influence of multi-level wavelet on the model performance, we designed a series of studies in TABLE~\ref{Wavelet}. Among them, cases 1, 2, and 3 denote the different levels of wavelet with fixed patch size. According to these results, we can find that when the level of wavelet increases, the required execution time and GPU memory consumption will be greatly reduced, but it cannot be ignored that the performance of the model will also decrease. This is because multiple downsampling operation makes the resolution of the image gradually decrease, so the GPU memory consumption is also greatly reduced. However, low-resolution will also cause the loss of local information of the image, making it difficult to reconstruct high-quality images. Therefore, multi-level wavelet-based models can be applied to mobile devices, which have strict restrictions on memory and execution time. In summary, the wavelet is effective to balance model performance and resource consumption. At the same time, multi-level wavelet can be considered according to actual needs.

\begin{table}[t]
\small
 \centering
 \setlength{\tabcolsep}{0.8mm}
   \caption{Study of DFEB's branch strategy (Kodak24, $\sigma$ = 30, color).}
  \begin{tabular}{@{}c|ccccccc@{}}
   \toprule
   Case        &Branch1     &Branch2       &Params     &Time       &GPU     &FLOPs      &PSNR            \\ 
   \midrule
   $1$       &Conv           &Conv           &6.45M      &2.44s      &1459MiB       &13.32G     &31.12dB              \\
   $2$       &Trans       &Trans          &6.08M   &7.37s  &6050MiB          &8.29G      &31.66dB              \\
   $3$      &Conv           &Trans          &6.12M  &6.21s  &4934MiB          &9.52G  &\textbf{31.72dB}     \\ 
   \bottomrule 
  \end{tabular}
  \label{Ablation1}
\end{table}

\begin{table}[t]
\centering
\small
\setlength{\tabcolsep}{2.2mm}
\caption{Study of the DFEB number to model performance (Kodak24, $\sigma$ = 30, color). xN stands for N DFEBs.}
\begin{tabular}{@{}ccccccc@{}}
			\toprule
			Case   	    &DFEBs    &Params     &Time       &GPU     &Flops      &PSNR	      	        \\ 
			\midrule
			$1$      	&$\times1$      &3.34M      &3.01s      &2656MiB          &5.44G        &31.55         \\
			$2$         &$\times2$      &6.12M      &6.21s      &4934MiB          &9.52G        &31.72                  \\
			$3$         &$\times3$      &8.84M      &9.84s      &6324MiB          &13.69G        &31.75                  \\
			$4$     	&$\times4$  	&11.8M      &11.96s		&7636MiB          &17.82G         &\textbf{31.78}                  \\ 
			\bottomrule  
\end{tabular}
\label{Ablation2}
\end{table}


\subsection{DFEB Investigations} 
As the most important part of EWT, Dual-stream Feature Extraction Block (DFEB) is designed for feature extraction while reducing the model size and shortening the running time. This is benefits from the double-branch structure in DFEB, which can elegantly combine CNN and Transformer. In order to verify the effectiveness of this strategy, we designed a series of experiments in TABLE~\ref{Ablation1}. Among them, all models only use two DFEBs and are trained with patchsize=64 for quick verification. According to the table, we can observe that the use of convolutional layers will lead to an increase in the number of parameters and FLOPs, and the use of Transformer will lead to more GPU memory consumption and longer execution time. Therefore, the model using our proposed strategy achieves intermediate results across multiple metrics. However, it is worth mentioning that our method achieves the best PSNR result and has a good balance between the performance, execution time, GPU memory consumption, FLOPs, and size of the model. All these results fully validate the necessity and effectiveness of the combination of CNN and Transformer.

In addition to this, we also study the impact of the number of DFEBs on model performance, execution time, and GPU usage in TABLE~\ref{Ablation2}. In this part, we set the patchsize to 64 to speed up training. Obviously, when the number of DFEBs is increased from 1 to 2, the model performance improves by 0.17dB. Continuing to increase the number of DFEBs can further improve the performance of the model, but the growth rate will gradually decrease. At the same time, it cannot be ignored that as the number of DFEBs increases, the GPU memory consumption and execution time of the model will greatly increase. Therefore, to ensure the efficiency of the model, we use 4 DFEBs in the final version of EWT.

\begin{table}[!t]
\small
\centering
\setlength{\tabcolsep}{0.7mm}
\caption{Detailed comparison study with Transformer method under Gaussian noise condition (noise level $\sigma$ = 30).}
\begin{tabular}{c|c|c|c|c|c}
\toprule
Method                      & GPU                      & Params             & Dataset    & PSNR             &  Time \\
\midrule
\multirow{3}{*}{SwinIR~\cite{r19}}    & \multirow{3}{*}{18432MiB}         & \multirow{3}{*}{5.17M} & Kodak24 & \textbf{31.79dB}          & 53.29s         \\
                            &                                 &                        & CBSD68  & \textbf{30.64dB} & 85.91s         \\
                            &                                 &                        & CUrban100  & \textbf{31.36dB} & 232.46s         \\
\midrule
\multirow{3}{*}{Uformer~\cite{wang2022uformer}}    & \multirow{3}{*}{6875MiB}         & \multirow{3}{*}{5.28M} & Kodak24 & 31.57dB          & 9.46s         \\
                            &                                 &                        & CBSD68  & 30.07dB & 16.21s         \\
                            &                                 &                        & CUrban100  & 30.82dB & 44.50s         \\
\midrule
\multirow{3}{*}{Restormer~\cite{zamir2022restormer}}    & \multirow{3}{*}{21894MiB}         & \multirow{3}{*}{12.47M} & Kodak24 & 31.62dB          & 42.86s         \\
                            &                                 &                        & CBSD68  & 30.51dB  & 82.53s         \\
                            &                                 &                        & CUrban100  & 31.16dB & 215.02s         \\
\midrule
\multirow{3}{*}{EWT} & \multirow{3}{*}{\textbf{6347MiB}} & \multirow{3}{*}{5.18M} & Kodak24 & 31.73dB          & \textbf{9.14s} \\
                            &                                 &                        & CBSD68  & 30.60dB          & \textbf{14.34s}  \\   
                            &                                 &                        & CUrban100  & 31.35dB & \textbf{43.77s}         \\
\bottomrule 
\end{tabular}
\label{Swin}
\end{table}

\subsection{Comparision with SwinIR}
In the previous subsection, we compared EWT with SwinIR~\cite{r19} to verify the positive effect of wavelet on the model. Here we provide more datasets and methods (Uformer~\cite{wang2022uformer} and Restormer~\cite{zamir2022restormer} ) to further verify the
effectiveness of EWT. All models are retrained under the same dataset and training settings. In TABLE~\ref{Swin} we provide the number of parameters
of the model, GPU memory used for training, PSNR results and average execution time on different test sets. As can be seen from the results in the table, EWT achieved better results than Uformer and Restorer with less GPU memory and
execution time, maintaining a good balance between performance and operating efficiency. It is worth noting that Uformer does improve efficiency through multi-level downsampling but seriously affects the performance of the model. This is why we introduced the wavelet transform to replace the downsampling
operation since the downsampling operation will cause a large number of features to be lost. On the whole, our EWT is a very potential method for image denoising and provide a new solution for image restoration.


\begin{figure}[h]
	\centering
	\begin{minipage}[c]{0.155\textwidth}
		\includegraphics[width=2.7cm,height=2cm]{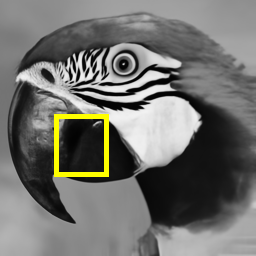}
		\includegraphics[width=2.7cm,height=2cm]{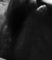}
		 \centerline{SwinIR}
		 \centerline{26.91dB}
	\end{minipage}
    \begin{minipage}[c]{0.155\textwidth}
		\includegraphics[width=2.7cm,height=2cm]{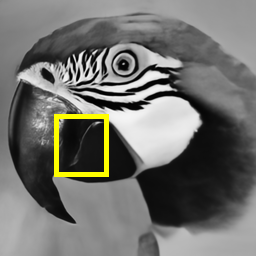}
		\includegraphics[width=2.7cm,height=2cm]{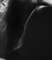}
		\centerline{EWT (Ours)}
		\centerline{26.82dB}
	\end{minipage}
	\begin{minipage}[c]{0.155\textwidth}
		\includegraphics[width=2.7cm,height=2cm]{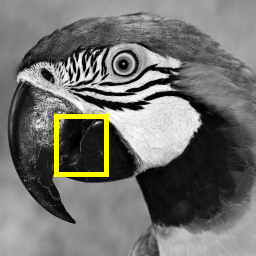}
		\includegraphics[width=2.7cm,height=2cm]{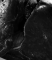}
		\centerline{GT}
		\centerline{PSNR}
	\end{minipage}
	\caption{Visual comparison with SwinIR~\cite{r19} on grayscale image. Obviously, EWT can reconstruct more accurate and clear edges. (Set12~\cite{zeyde2010single}, $\sigma$ = 50)}
	\label{Visual-2}
\end{figure}

\begin{figure}[h]
	\centering
		\begin{minipage}[c]{0.155\textwidth}
		\includegraphics[width=2.7cm]{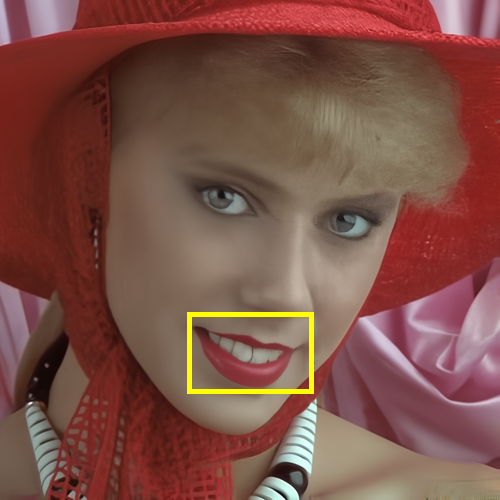}
		\includegraphics[width=2.7cm]{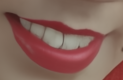}
		 \centerline{29.49dB}
	\end{minipage}
    \begin{minipage}[c]{0.155\textwidth}
		\includegraphics[width=2.7cm]{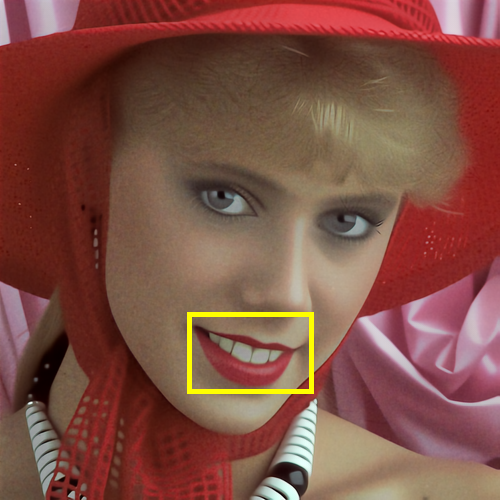}
		\includegraphics[width=2.7cm]{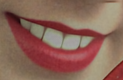}
		\centerline{29.19dB}
	\end{minipage}
	\begin{minipage}[c]{0.155\textwidth}
		\includegraphics[width=2.7cm]{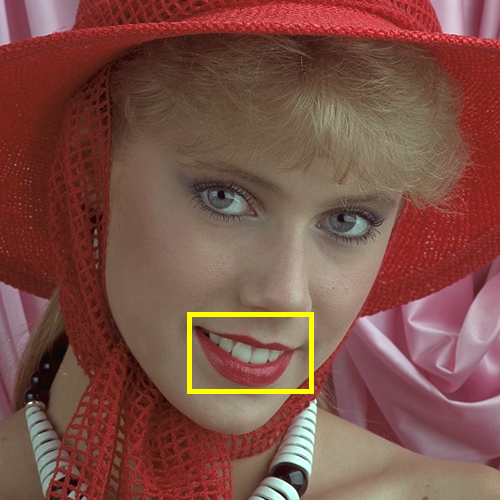}
		\includegraphics[width=2.7cm]{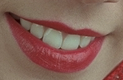}
		\centerline{PSNR}
	\end{minipage}
	
	\begin{minipage}[c]{0.155\textwidth}
		\includegraphics[width=2.7cm,height=1.65cm]{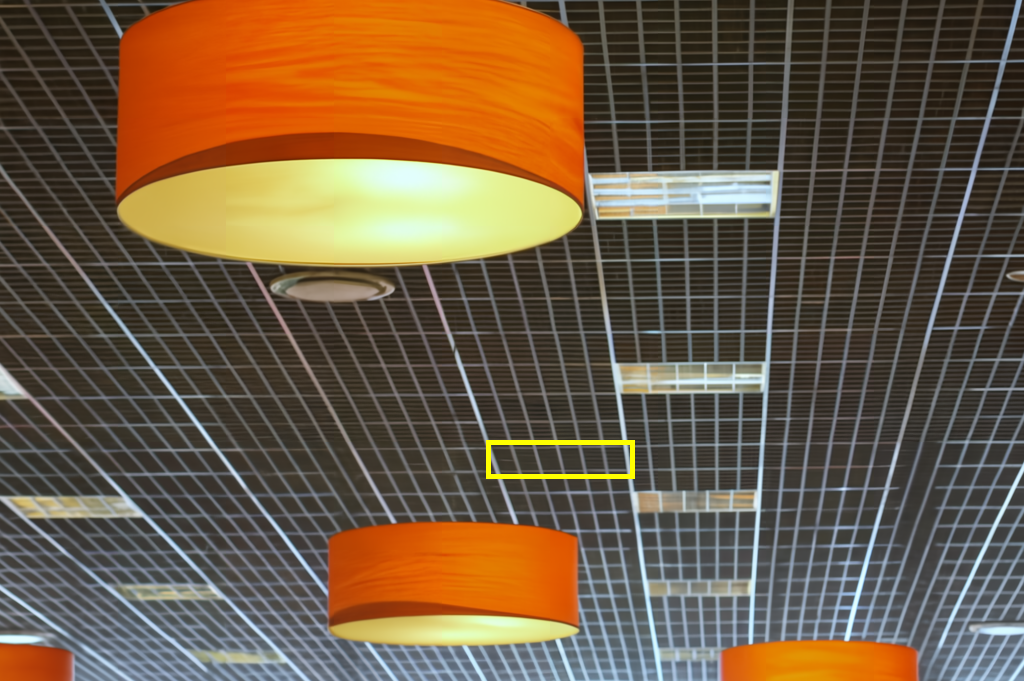}
		\includegraphics[width=2.7cm,height=1.65cm]{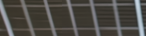}
		 \centerline{33.16dB}
		 \centerline{SwinIR}
	\end{minipage}
    \begin{minipage}[c]{0.155\textwidth}
		\includegraphics[width=2.7cm,height=1.65cm]{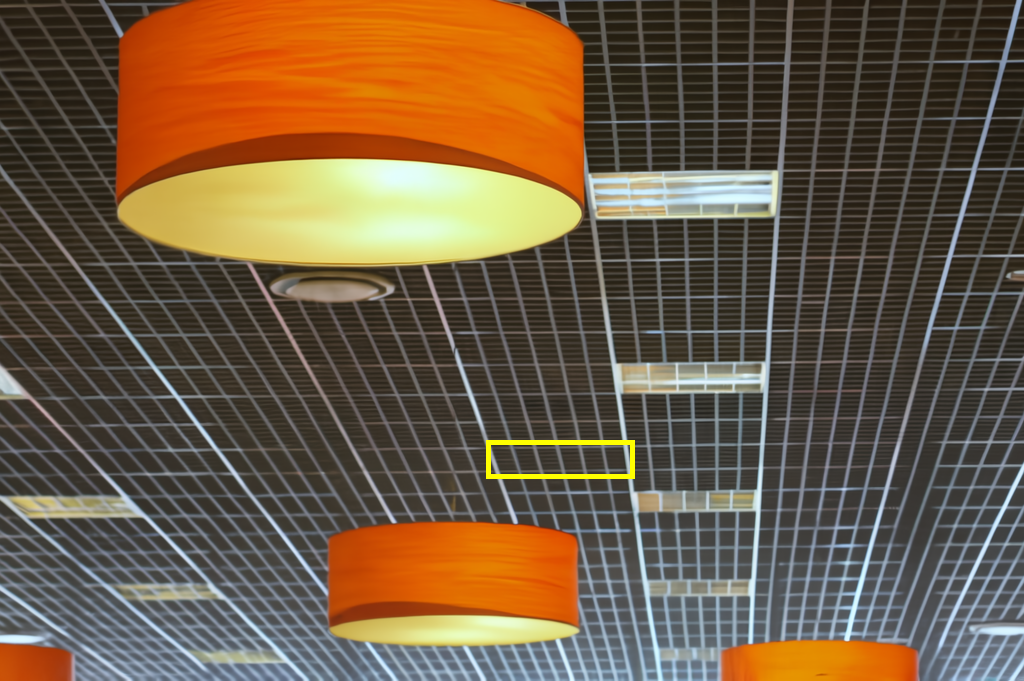}
		\includegraphics[width=2.7cm,height=1.65cm]{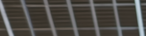}
		\centerline{32.93dB}
		\centerline{EWT (Ours)}
	\end{minipage}
	\begin{minipage}[c]{0.155\textwidth}
		\includegraphics[width=2.7cm,height=1.65cm]{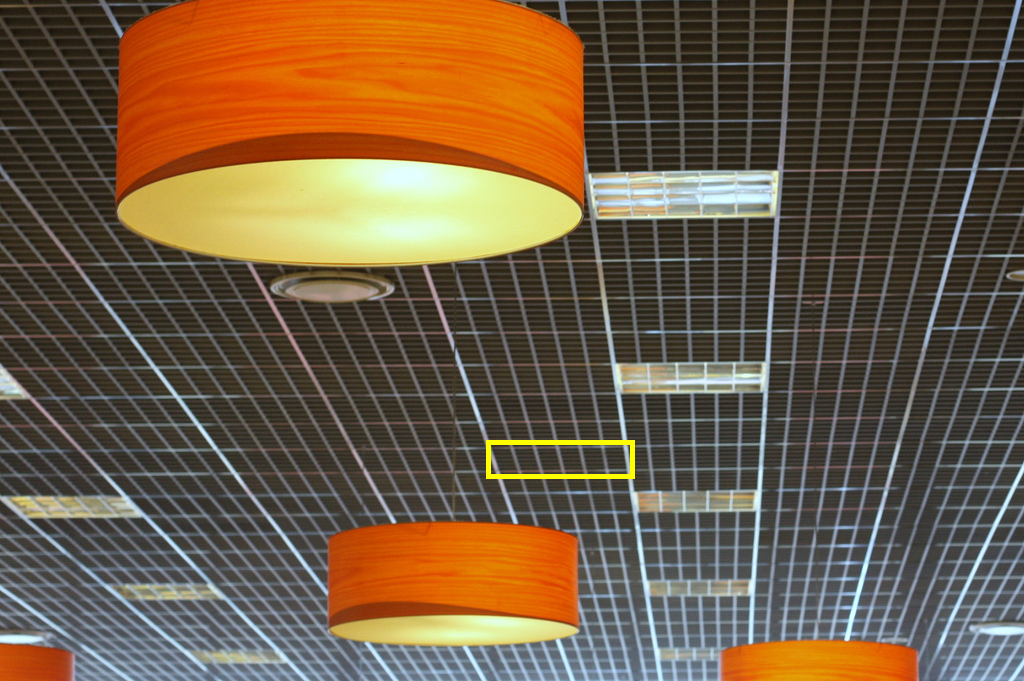}
		\includegraphics[width=2.7cm,height=1.65cm]{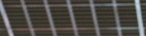}
		\centerline{PSNR}
		\centerline{GT}
	\end{minipage}
	\caption{Visual comparison with SwinIR~\cite{r19} on color image. Obviously, EWT can reconstruct more accurate and clear lines. ($\sigma$ = 50)}
	\label{Visual-1}
\end{figure}

In Figs.~\ref{Visual-2} and~\ref{Visual-1}, we provide the visual comparisons with SwinIR~\cite{r19} on grayscale and color images, respectively. It is worth noting that the SwinIR results used here are the denoised image reconstructed by the original paper provided pre-trained model, which uses DIV2K~\cite{DIV2K} (800 training images), Flickr2K~\cite{lugmayr2019aim} (2650 images), BSD500~\cite{martin2001database} (400 training and testing images) and Waterloo Exploration Database~\cite{ma2016waterloo} (4744 images) for training. However, our EWT only use 800 training images from DIV2K, which is 1/10 of the SwinIR training set. According to the results, we can clearly observe that although SwinIR achieved slightly better PSNR results than our EWT, the reconstructed denoised images are also smoother and lack texture details. In contrast, our EWT can reconstruct sharper and more accurate image edges. This is because the introduced wavelet can capture the frequency and position information of the image, which is beneficial to restore the detailed features of the image. Therefore, we can draw the following conclusions: (1). Compared with SwinIR, our EWT can achieve close results with less GPU memory consumption and faster inference time; (2). Compared with SwinIR, our reconstructed denoised images have richer texture details and more accurate edges. All these results further validate the effectiveness of EWT. To sum up, our method has more advantages than previous Transformer-based models, which achieve a good balance between the performance and efficiency of the model.

\subsection{Comparision with MWCNN}

In this paper, we proposed a novel Efficient Wavelet-Transformer (EWT) for single image denoising. This is the first attempt of Transformer in wavelet domain. As we mentioned in the previous section, EWT was proposed inspired by MWCNN~\cite{r12}. Therefore, we give a detailed comparison with MWCNN in TABLE~\ref{WMCNN}. According to the table, we can clearly observe that our EWT achieves better results on the vast majority of datasets and noise levels with fewer parameters. This fully demonstrates the effectiveness of the proposed EWT. Meanwhile, it also means that it is meaningful and feasible to combine wavelet and Transformer, which further promoted the development of the wavelet in SID.

\begin{table}[http]
\centering
\setlength{\tabcolsep}{3.0mm}
\caption{PSNR (dB) and parameter quantity comparison with DHDN~\cite{yu2019deep} and DIDN~\cite{park2019densely} on color image test datasets.}
\begin{tabular}{l|c|c|c|c}
\toprule
Method                  & Noise Level & DHDN           & DIDN  & EWT (Ours)     \\ 
\midrule
\multirow{3}{*}{Kodak24}  & $\sigma$=10         & \textbf{37.33dB}          & 37.32dB & 37.24dB          \\ \cline{2-5} 
                        & $\sigma$=30         & 31.95dB          & \textbf{31.97dB} & 31.90dB          \\ \cline{2-5} 
                        & $\sigma$=50         & 29.67dB & \textbf{29.72dB} & 29.63dB \\ \hline
\multirow{3}{*}{CBSD68} & $\sigma$=10         & 36.45dB          & 36.48dB & \textbf{36.49dB} \\ \cline{2-5} 
                        & $\sigma$=30         & 30.41dB          & 30.70dB & \textbf{30.71dB} \\ \cline{2-5} 
                        & $\sigma$=50         & 28.02dB          & 28.35dB & \textbf{28.36dB} \\ \hline
\multicolumn{2}{c|}{Parameters}                     & 168M           & 165M  & \textbf{11.8M}          \\ 
\bottomrule 
\end{tabular}
\label{Big}
\end{table}

\subsection{Model Size Investigations}

Increasing the depth of the model is the easiest way to improve the model performance. However, it cannot be ignored that these models~\cite{zhang2020residual,yu2019deep,park2019densely} also accompanied by a large number of parameters. In Fig.~\ref{Size}, we provide the performance and parameter comparisons of EWT with other SID models, including IRCNN~\cite{zhang2017learning}, DnCNN~\cite{r13}, FFDNet~\cite{zhang2018ffdnet}, ADNet~\cite{tian2020attention}, BRDNet~\cite{tian2020image}, MLEFGN~\cite{fang2020multilevel}, RNAN~\cite{zhang2019residual}, RDN~\cite{zhang2020residual}, DIDN~\cite{park2019densely}, and IPT~\cite{r18}. Among them, the red star represents EWT. Obviously, EWT achieves competitive results with few parameters, which strike a good balance between the performance and size of the model. Moreover, we provide a detailed comparison with DHDN~\cite{yu2019deep} and DIDN~\cite{park2019densely} in TABLE~\ref{Big}. \textbf{Obviously, EWT achieves best results on CBSD68 and close results on Kodak24 with only 1/14 parameters of DHDN and DIDN.} All these results validate that EWT is an efficient and accurate SID model.

\begin{table*}
\centering
\setlength{\tabcolsep}{2.7mm}
\caption{Comparison with MWCNN on grayscale images. The best results are \textbf{highlighted}.}
\begin{tabular}{l|c|ccc|ccc|ccc}
\toprule
Method      & Parameters & \multicolumn{3}{c}{Set12}               & \multicolumn{3}{c}{BSD68}                        & \multicolumn{3}{c}{Urban100}                     \\
\midrule
Noise Level &            & $\sigma$ =15             & $\sigma$ =25             & $\sigma$ =50    & $\sigma$ =15             & $\sigma$ =25             & $\sigma$ =50             & $\sigma$ =15             & $\sigma$ =25             & $\sigma$ =50             \\ 
\midrule
MWCNN~\cite{r12}       & 19.2M      & 33.15dB          & 30.79dB          & 27.74dB & 31.86dB          & \textbf{29.41dB} & \textbf{26.53dB} & 33.17dB          & 30.66dB          & 27.42dB          \\
EWT (Ours)  & 11.8M      & \textbf{33.23dB} & \textbf{30.86dB} & \textbf{27.80dB} & \textbf{31.87dB} & 29.40dB          & 26.47dB          & \textbf{33.54dB} & \textbf{31.08dB} & \textbf{27.70dB}\\
\bottomrule 
\end{tabular}
\label{WMCNN}
\end{table*}

\section{Discussion}\label{DS}

In this paper, we proposed an Efficient Wavelet-Transformer (EWT) and demonstrate its effectiveness on the SID task. However, this does not mean that it is only suitable for SID. EWT is a general model that can be applied to other image restoration tasks, such as image super-resolution, image dehazing, and image deraining. In future works, we will further explore its effectiveness on other image restoration tasks, and optimize the model according to different tasks.

\section{Conclusion}\label{CL}
In this paper, a novel Efficient Wavelet-Transformer (EWT) is proposed for single image denoising. Specifically, we introduced Discrete Wavelet Transform (DWT) and Inverse Wavelet Transform (IWT) for downsampling and upsampling operations, respectively. This method can greatly reduce the resolution of the image, thereby reducing GPU memory consumption, and will not cause any loss of information. Meanwhile, an efficient Multi-level Feature Aggregation Module (MFAM) is proposed to make full use of hierarchical features by using local and global residual learning. In addition, a novel Dual-stream Feature Extraction Block (DFEB) is specially designed for local and global features extraction, which combines the advantages of CNN and Transformer that can take into account the information of different levels. Extensive experiments show that our EWT achieves the best balance between the performance, size, execution time, and GPU memory consumption of the model.

\bibliographystyle{unsrt}

 




\vfill

\end{document}